# ImJoy: an open-source computational platform for the deep learning era

Wei Ouyang[1,2]*, Florian Mueller[1]*, Martin Hjelmare[2], Emma Lundberg[2,3,4], Christophe Zimmer[1]

[1] Imaging and Modeling Unit, Institut Pasteur, UMR 3691 CNRS, C3BI USR 3756 IP CNRS, Paris, France

[2] Science for Life Laboratory, School of Engineering Sciences in Chemistry, Biotechnology and Health, KTH - Royal Institute of Technology, Stockholm, Sweden

[3] Department of Genetics, Stanford University, Stanford, CA 94305, USA (visiting appointment)

[4] Chan Zuckerberg Biohub, San Francisco, CA 94158, USA (visiting appointment)

* To whom correspondence should be addressed.

Wei Ouyang: weio@kth.se

Florian Mueller: muellerf.research@gmail.com

**Deep learning methods have shown extraordinary potential for analyzing very diverse biomedical data, but their dissemination beyond developers is hindered by important computational hurdles. We introduce ImJoy (https://imjoy.io/), a flexible and open-source browser-based platform designed to facilitate widespread reuse of deep learning solutions in biomedical research. We highlight ImJoy's main features and illustrate its functionalities with deep learning plugins for mobile and interactive image analysis and genomics.**

Deep learning methods, which use artificial neural networks to learn complex mappings between numerical data, have enabled recent breakthroughs in a wide range of biomedical data analysis tasks. Examples for imaging data include image segmentation[1,2] and medical diagnosis, where deep learning vastly outperforms more traditional methods and often exceeds human expert performance[3,4], or methods to enhance microscopy images, e.g. for denoising or super-resolution, or to predict fluorescence-like channels from images of unstained samples[5–8]. In genomics, several studies also successfully use deep learning to analyze DNA sequence data, for instance to predict the specificity of DNA-binding proteins or the effects of non-coding variants[9,10]. Many other promising applications abound, for example in protein structure determination, chemical synthesis and drug design[11,12].

These and other early advances generate considerable interest and a strong demand by biomedical researchers to apply and adapt deep learning methods to new data sets and questions. The potential of deep learning is further increased by the growing number of public data repositories in genomics, imaging and other fields[13,14]. However, making full use of recent deep learning approaches faces considerable bottlenecks. A distinctive challenge of machine learning methods arises from their strong reliance on training data. Published studies occasionally provide already trained neural networks (models), sometimes in the form of easy-to-use tools, such as web applications or ImageJ plugins[2,5,6,10]. While these tools are useful, they do not allow researchers to retrain models on their own data or on public data sets. This restriction severely limits the reuse of deep learning methods, because models trained on one data set do not necessarily perform well on different types of data (model mismatch), potentially resulting in artifacts[15], and as a general rule much better results can be achieved by full or partial retraining on new or larger data sets. Therefore, enabling users to retrain existing models on other data is essential to realizing the full promise of deep learning in biomedical research.

Many deep learning approaches provide open-source code, typically written in Python and using libraries such as Tensorflow or Pytorch, which in theory enables retraining. In practice, however, setting up the required computational environments, which involve many hardware and operating system-dependent libraries with complex dependencies, can be daunting. Furthermore, although these implementations allow prototyping by developers, they use command line instructions and are thus difficult to employ for researchers lacking adequate programming skills. Cloud computing services can alleviate some of these difficulties[1], but require users to upload their data, which can cause privacy and confidentiality issues, or conflict with regulations such as GDPR, especially for biomedical data[16]. Thus, empowering biomedical researchers to take full advantage of recent advances in deep learning currently remains an important challenge.

To address this challenge, we developed ImJoy (https://imjoy.io), an open-source platform designed to deliver advanced, yet easy-to-use data analysis tools, especially based on deep learning. The core of ImJoy is a browser based application (i.e. a software running in a web browser) that offers a rich and interactive user interface and leverages modern web technologies (**Fig 1a, Supplementary Figure 1**). ImJoy's data analysis functionalities are provided by a flexible system of plugins that operate independently of each other but can be organized into workspaces and workflows (**Supplementary Figure 2**). Depending on the need, ImJoy can access various computational resources by performing computations either in the browser itself or using 'plugin engines' that can run either locally or remotely (**Fig 1a, Supplementary Figure 1**). Below, we highlight ImJoy's main features, along with illustrative plugins with installation links and descriptions being provided in the Methods. A detailed documentation on how to use ImJoy and its plugins and how to develop new ones is provided online (https://imjoy.io/docs).

ImJoy provides self-contained plugin development support with a built-in code editor for developing and testing plugins. It permits the use of plugins written in different programming languages (JavaScript, HTML, and Python) within the same analysis workflow. ImJoy's API enables bidirectional data exchange and allows plugins to use functions from other plugins (**Supplementary Figure 2)**. Plugins are stored as separate files, can be hosted on GitHub, and distributed with a single link, e.g. by email or social media (**Fig 1b**). This link directly opens ImJoy and, importantly, allows developers to control and automate complex installations, e.g. to set up platform-dependent deep learning libraries (see Methods for more details). This makes installation transparent to the user and removes a major obstacle for tool deployment and reuse by researchers. ImJoy also makes it easy to develop plugins that leverage a large variety of available advanced user-interface libraries. As a result, ImJoy plugins can be optimized for responsiveness and flexibility, making them suitable for different screen sizes including smartphones. Furthermore, these libraries provide vast possibilities for interactive data-visualization. As an illustration, we implemented a plugin for manual annotation of structures of interest in images, a key requirement for defining training targets in image segmentation (**Methods**: Image Annotator plugin). Another example plugin uses uses high dimensional features previously extracted from deep learning models to analyze the localization of proteins based on data from the Human Protein Atlas[17] (HPA, https://proteinatlas.org, **Methods**: HPA-UMAP plugin, **Fig 2a**). Using this plugin, researchers can visualize this entire data set as scatter plots organized by feature-similarity (using dimensionality reduction techniques) and interactively inspect the raw images associated to individual data points. Such interactive visualization capabilities will be instrumental to explore large image databases.

ImJoy offers flexible access to multiple resources for computational analysis, which can be adapted to the computational requirements of each task. For many applications, computations can be performed efficiently within the browser itself. Indeed, beyond their original purpose of navigating the Internet, web browsers are now increasingly used as computational platforms. Thanks to sandboxing inherent to browser computing, ImJoy also offers higher levels of security and stability than non-browser based software. ImJoy supports several browser-based libraries that offer advanced data processing capabilities. For example, TensorFlow.js allows to run deep learning models on graphical processing units (GPUs), while WebAssembly allows to run Python code in the browser. Note that browser-based computing does not require an Internet connection. As an example, we implemented a plugin to classify images of potentially cancerous skin lesions[18] using deep convolutional networks (**Methods**: Skin-Lesion-Analyzer plugin). Thanks to the interface versatility mentioned above, the plugin can run on a smartphone and directly analyze skin images taken with the camera (**Fig 1b**). Importantly, the actual classification task is performed in the phone's browser without sending any data over the Internet, thus mitigating privacy concerns. Another example is a plugin to classify protein localization patterns by deep learning (**Methods**: HPA-classification plugin). In addition to predicting localization classes from images, this plugin also shows a class activation map[19], which highlights the image regions most important for the classification. Such information is valuable not only for developers aiming to fine-tune the algorithm and improve classification performance, but also for users seeking to interpret the output, e.g. for medical diagnosis[20].

To perform more demanding or computationally intensive tasks, such as training or retraining a deep net on large data sets, browser-based computation may not always suffice. Therefore, ImJoy can connect to a plugin engine (**Fig 1a, Supplementary Figure 1**), which can be launched on different hardware, ranging from a local computer to private servers or public cloud services. The plugin engine has access to the entire Python ecosystem, and uses Conda to automatically solve plugin dependencies and set up virtual environments to avoid dependency conflicts. This relieves users from setting up complex software dependencies and provides reproducible analysis workflows, which is crucial for data science[21]. Importantly, the unified plugin interface of ImJoy ensures that remote procedure calls to the plugin engine are transparent, i.e. the user will be exposed to the same interface independently of where computations actually occur. As an illustration, we implemented two ImJoy plugins for deep learning based image transformation methods: a plugin implements ANNA-PALM, a recent method to create dense super-resolution images from sparse localization data using neural networks[6] (**Methods**: ANNA-PALM plugin), and another implements CARE, which uses deep learning for 3D image restoration[5] (**Methods**: CARE plugin, **Fig 2a**). Other machine learning applications often require annotations performed by an expert before training, a task that is greatly facilitated by ImJoy's afore-mentioned user-interactivity. As an example, we implemented a deep learning nuclei segmentation framework. This framework is the winning entry on the 2018 Kaggle Data Science Bowl for nuclear sementation and based on the popular U-net architecture[2] (**Methods**: DPNUnet plugin). From a central plugin, the user can choose between plugins for annotation, mask generation, training, or prediction (**Fig 2b**). The annotation plugin  allows researchers to manually outline nuclei, using a computer mouse or touch-screen device, these annotations can then be used for mask generation. The training plugin trains the U-net on the local or remote plugin engine and the prediction plugin uses the trained U-net to segment new images.  For all these examples, training can be monitored in real time, e.g. by inspecting the loss function or outputs for individual samples (**Fig 2a**), which allows to fine-tune the algorithm - for example by adding regularization or perform early stopping to remedy overfitting. Together with appropriate computational resources (typically GPUs), ImJoy thus enables users to train these powerful methods on their own data or on other public data sets, which was not easily possible before.

Finally, we note that ImJoy is ideally suited to integrate with online databases and model repositories, providing a versatile front-end to efficiently leverage large numbers of data samples and methods. In addition to the HPA-UMAP plugin mentioned above, which already illustrated interactive exploration of a large image database, we also provide a plugin that implements DeepBind[10], a deep learning method to predict the sequence-specific binding affinities of proteins (**Methods**: DeepBindScan plugin, **Fig 2b**). This plugin is based on Kipoi.org - a collaborative initiative that provides ready-to-use trained models for genomics[22]. Unlike the original website and command line tool[10], this plugin allows to analyze any user-defined DNA or RNA sequence from a graphical user interface. Thousands of other models available on Kipoi.org can now easily be implemented through ImJoy.

We presented ImJoy, a new computational platform that permits the development of powerful data processing tools with rich and interactive interfaces, and allows computations to be performed in the browser, on local machines or devices, and remote servers. We highlighted plugins that illustrate interactive visualization capabilities, the use of pre-trained deep learning models, operation on smartphones, retraining of deep models on user-provided data, real-time monitoring of learning, applications to imaging and genomics data, and analysis of large public data repositories. ImJoy enables flexible hybrid computation modes, where analysis tasks can be distributed to a combination of local and remote computational environments. An interesting application is the processing of confidential medical data: raw images could be pre-processed and anonymised locally, for example to calculate feature vectors which could then be sent to a remote server for further analysis, e.g for comparison against a database. Although ImJoy was developed with a focus on deep learning, it can be used to implement and deploy a much broader range of computational methods. We foresee several extensions of ImJoy beyond its current implementation and plugins. Among these are the possibility to incorporate scripts already developed for other widely used data processing platforms, for example ImageJ/Fiji [22], allowing ImJoy to benefit from their functionalities and, conversely, confer new capabilities to these tools, such as easy deployment in the cloud. We also envision plugins that leverage hybrid computing to provide real-time feedback of freshly trained models to the user in response to annotations or other interactions. Such interactivity (human-in-the-loop) can result in better fine-tuning of the algorithm and improved learning[23,24].

In sum, we believe that ImJoy will greatly accelerate the adoption of deep learning technique in biomedical research by reducing the gap between developers and researchers and thereby open the door to new scientific discoveries, while also promoting reproducible research and open science.

## Acknowledgements

This work was funded by Institut Pasteur. W.O. was a scholar in the Pasteur–Paris University (PPU) International PhD program and was partly funded by a Fondation de la Recherche Médicale grant to C.Z. (DEQ 20150331762). W.O. is a postdoctoral researcher supported by the Knut and Alice Wallenberg Foundation (2016.0204) and Swedish Research Council (2017-05327). We also acknowledge Investissement d'Avenir grant ANR-16-CONV-0005 for funding a GPU farm used for testing ImJoy. We thank the IT department of Institut Pasteur, in particular Stéphane Fournier and Thomas Menard, for providing access to the kubernetes cluster and DGX-1 server for running and testing the ImJoy plugin engine and for technical support. We thank Quang tru Huynh for maintaining the GPU farm and for advice and assistance during the development of ImJoy. We also thank Anna Martinez Casals, Peter Thul, Hao Xu, Andrey Aristov, Anthony Cesnik, Christian Gnann, Alex Hu, Jyotsana Parmar, Kyle M. Douglass, Nico Stuurman, Xian Hao, Shubin Dai, David Guo, Kyrie Zhou for testing and helping with the ImJoy plugin development. We thank Elena Rensen for proofreading the manuscript. We thank Juan Nunez-Iglesias, Shalin Mehta, Bryant Chhun, Joshua Batson, Loic Royer, Nicholas Sofroniew, and Maxime Woringer for useful advice and discussion.

## Code availability

ImJoy (https://ImJoy.io) is open-source software released under MIT license. Code is hosted on GitHub: https://github.com/oeway/ImJoy.

17th April 2019)

## Figure Legends

**Figure 1**. **Overview of ImJoy.**

**a)** The ImJoy core is a browser-app. Plugins, specified libraries and other required resources (such as software libraries and pre-trained machine learning models) are retrieved from the Internet. Once these are downloaded and installed, ImJoy can run offline with a dedicated desktop app. ImJoy permits computing either in the browser, or can connect to multiple plugin engines, which can perform computations locally on a workstation, computing cluster, or remote servers and cloud services. **b)** Developers can write and test plugins directly in ImJoy, then deploy them as single plugin files to a GitHub repository, which can then serve as a plugin store for ImJoy. Plugins can be shared with a hyperlink that permits their installation and usage on different devices including smartphones.

**Figure 2**. **ImJoy example plugins**.

**a)** The "HPA-UMAP" plugin allows interactive exploration of images from the HPA cell atlas. The scatter plot shows a UMAP representation of features computed by a deep model from images of fluorescently tagged proteins[26]. Each dot corresponds to an image. The deep model has been trained to classify localization patterns, therefore dots that are close together in the UMAP tend to have similar localization patterns. When the user clicks on an individual dot, the plugin displays the corresponding image on the right. **b)** The ImJoy Dashboard allows users to monitor training of a deep model in real-time. Top: the learning curve plots the loss, which measures the model's performance on validation data, as function of iterations. Bottom: the user can inspect predictions on individual samples during training. Here, the three images show the noisy input (left), the output of the currently trained CARE model[5] (middle), and the training target (i.e.

ground-truth) image (right). **c)** The DPNUnet plugin provides access to a complete framework for segmentation of nuclei based on the winning entry of the 2018 Kaggle data challenge. Images can be annotated with a dedicated plugin, allowing to generate masks for training. The pre-trained model can be retrained on these data and applied for segmentation of new images using two other plugins. **d)** The "DeepBindScan" plugin uses all available DeepBind[10] models to predict protein binding affinities against a user defined DNA- or RNA sequence.

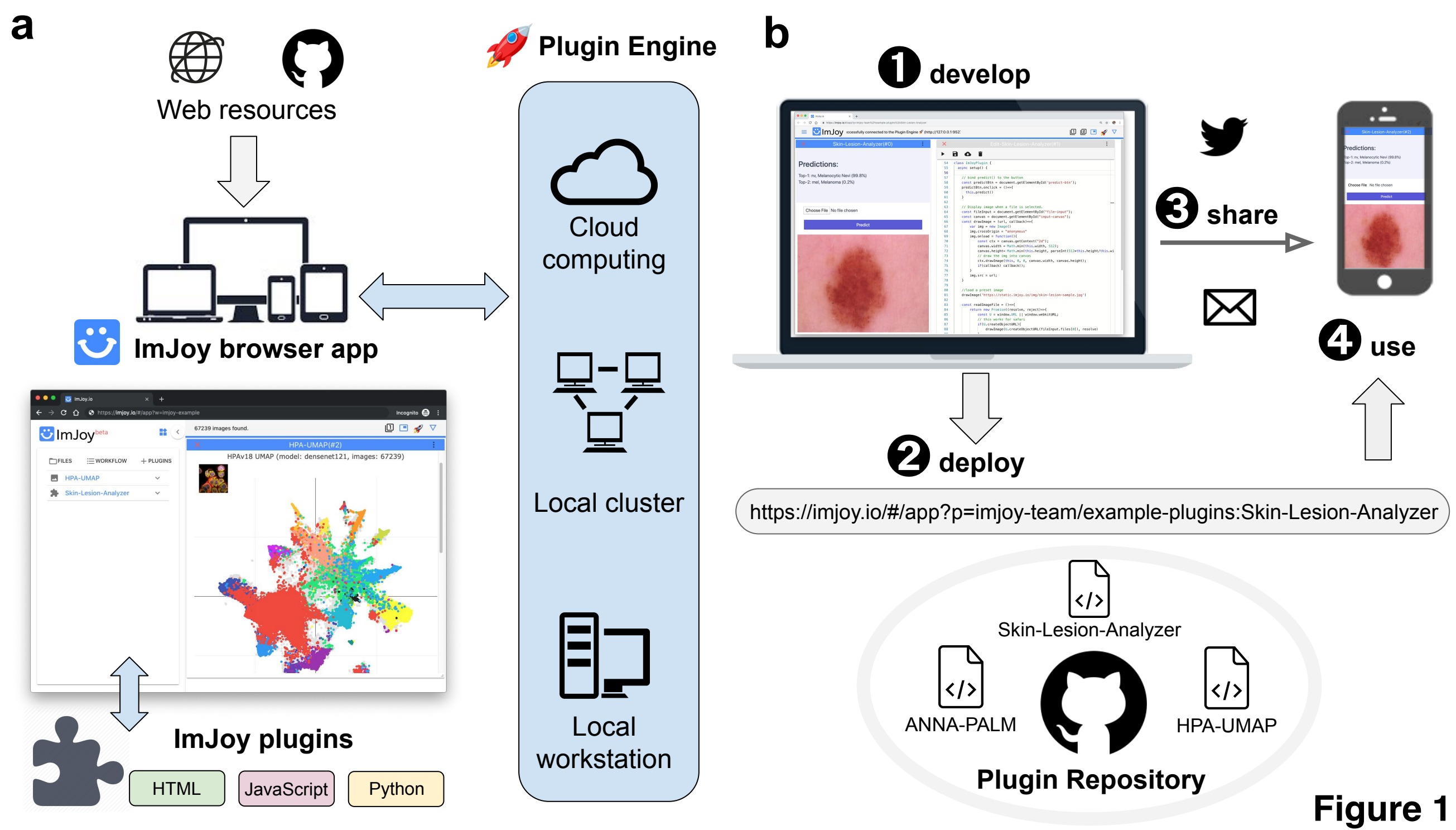

**Figure 1**

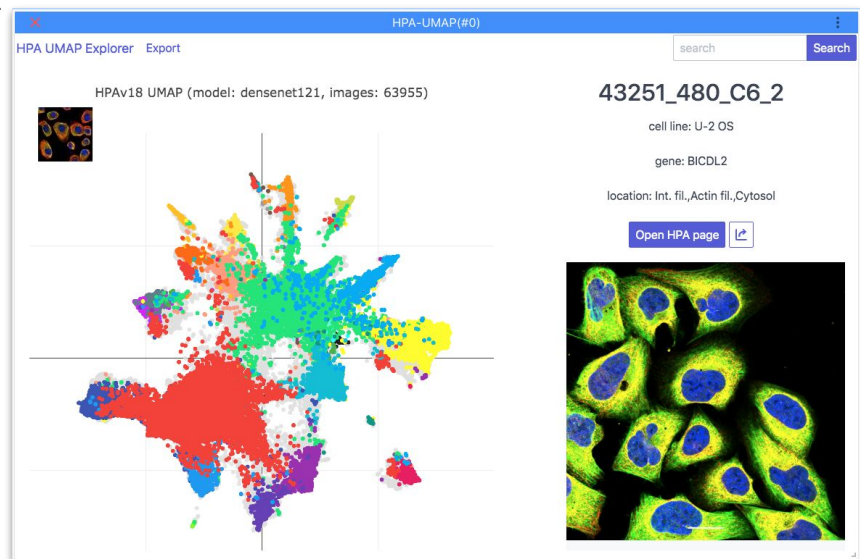

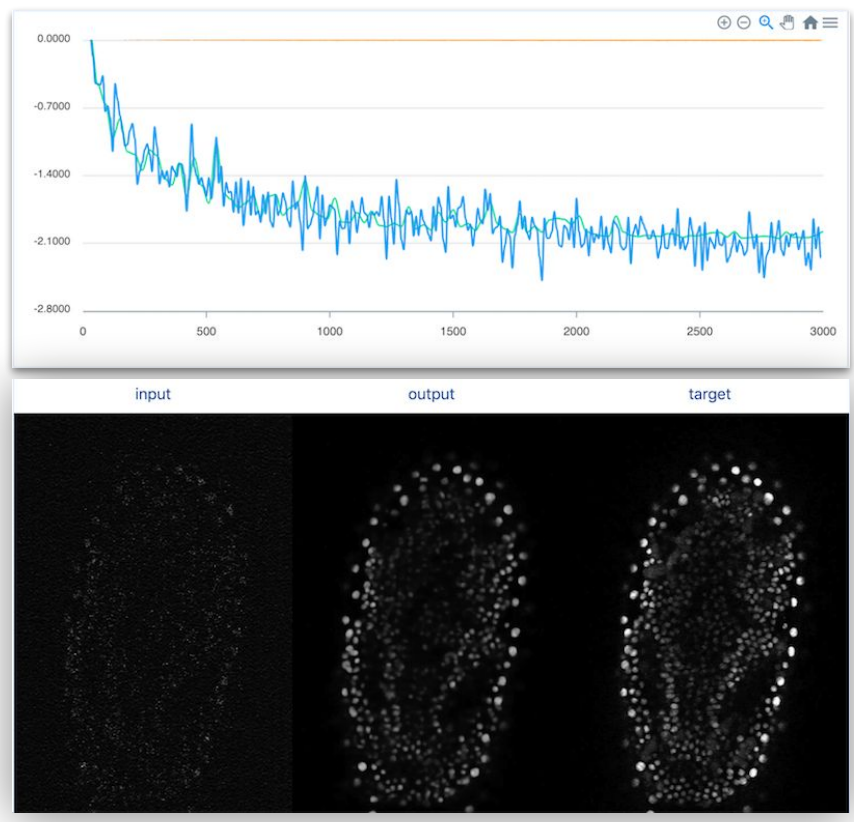

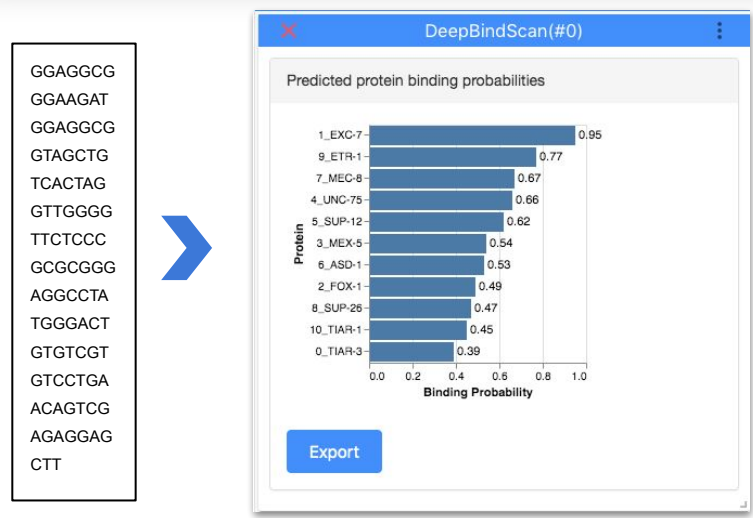

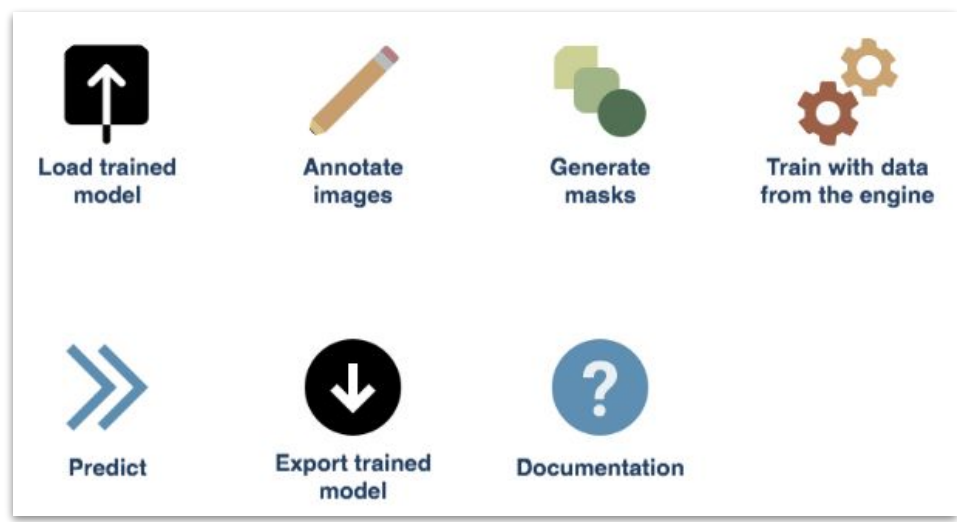

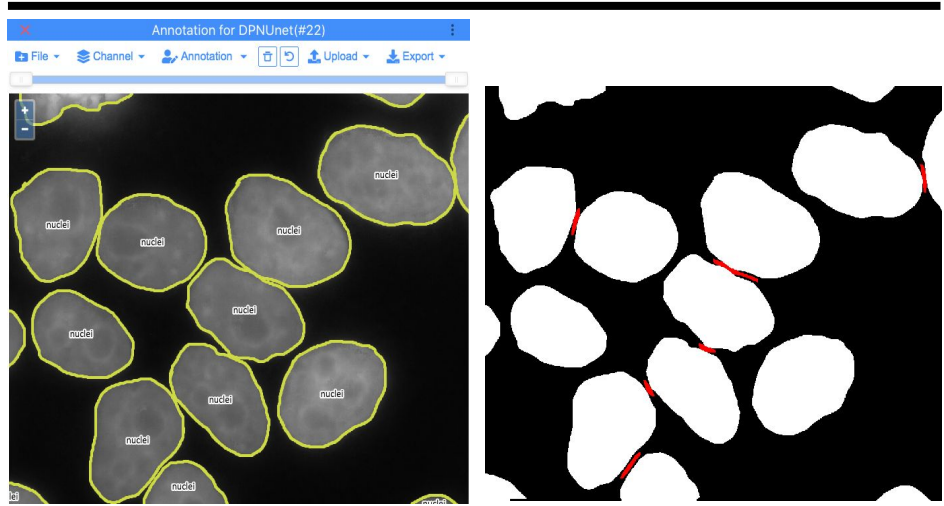

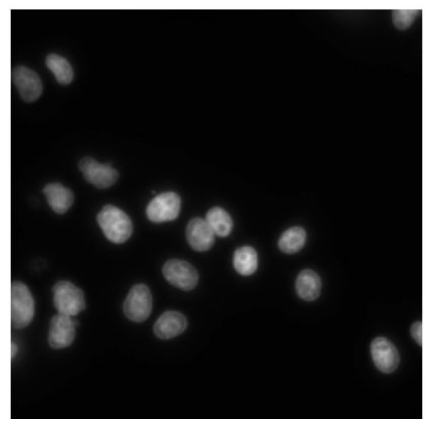 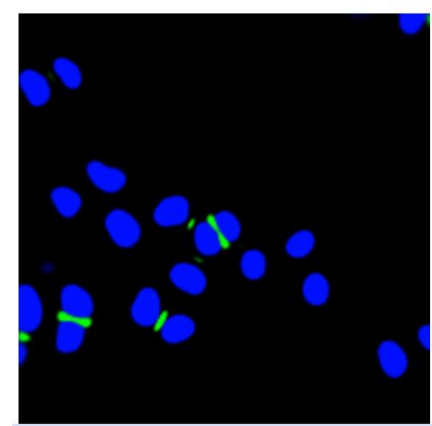

**Figure 2**

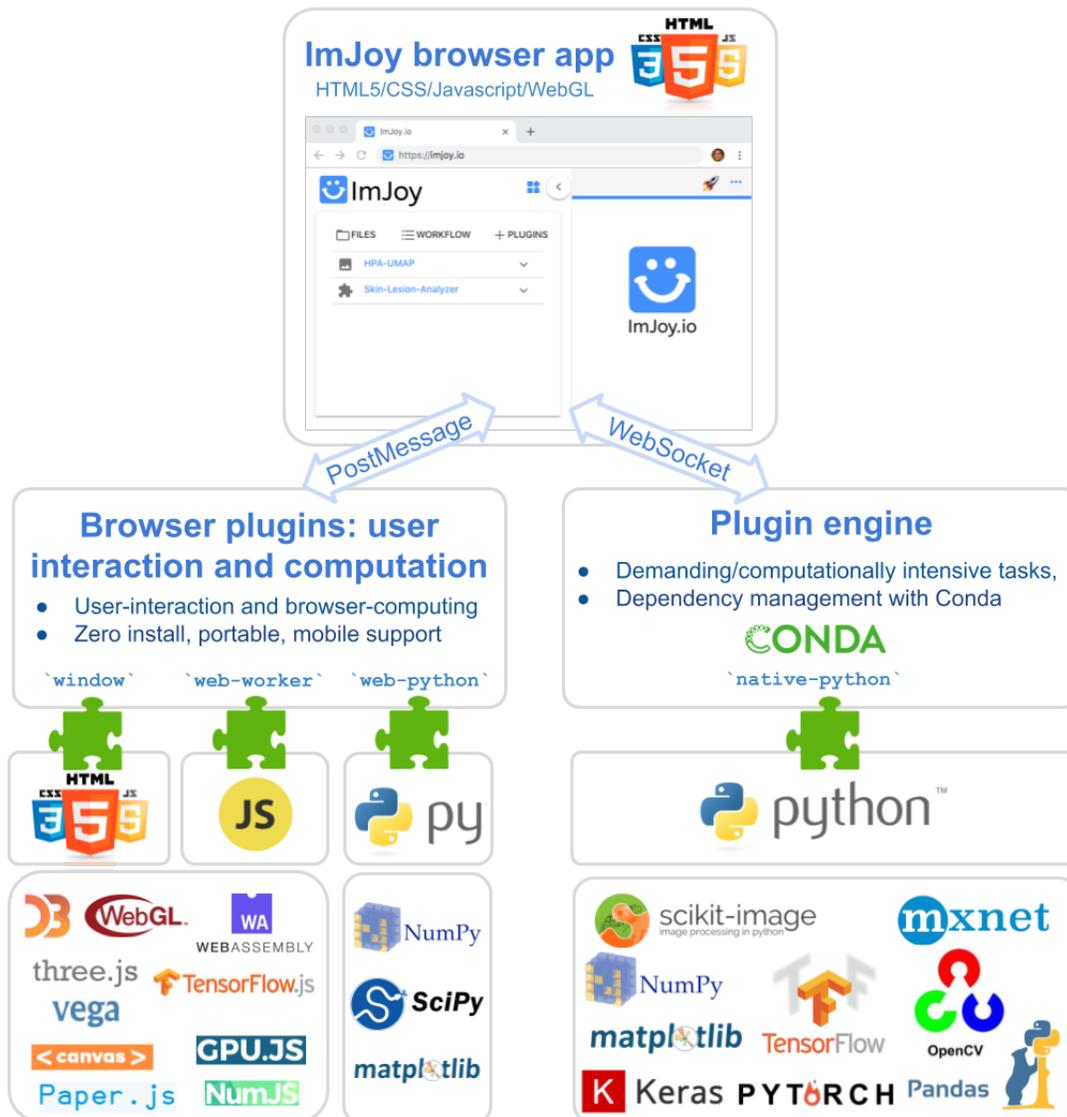

**Supplementary Figure 1: Overview of ImJoy architecture**.

ImJoy is a browser application developed with modern web technologies (top). Its functionalities are implemented with plugins that can run either directly in the browser or on a dedicated plugin engine (bottom). Browser plugins allow interacting with the user, visualizing data with a large variety of web libraries, performing computational tasks in webworkers, which enable computational tasks to run in a dedicated thread, or using Python through WebAssembly. Native Python plugins are supported through plugin engines implemented in Python and have access to the entire Python ecosystem, allowing them to perform demanding computational tasks with powerful scientific computing libraries such as Numpy, Scipy, or Scikit-image and to leverage deep learning tools such as Tensorflow and Pytorch. Virtual environments and dependent packages are managed with Conda.

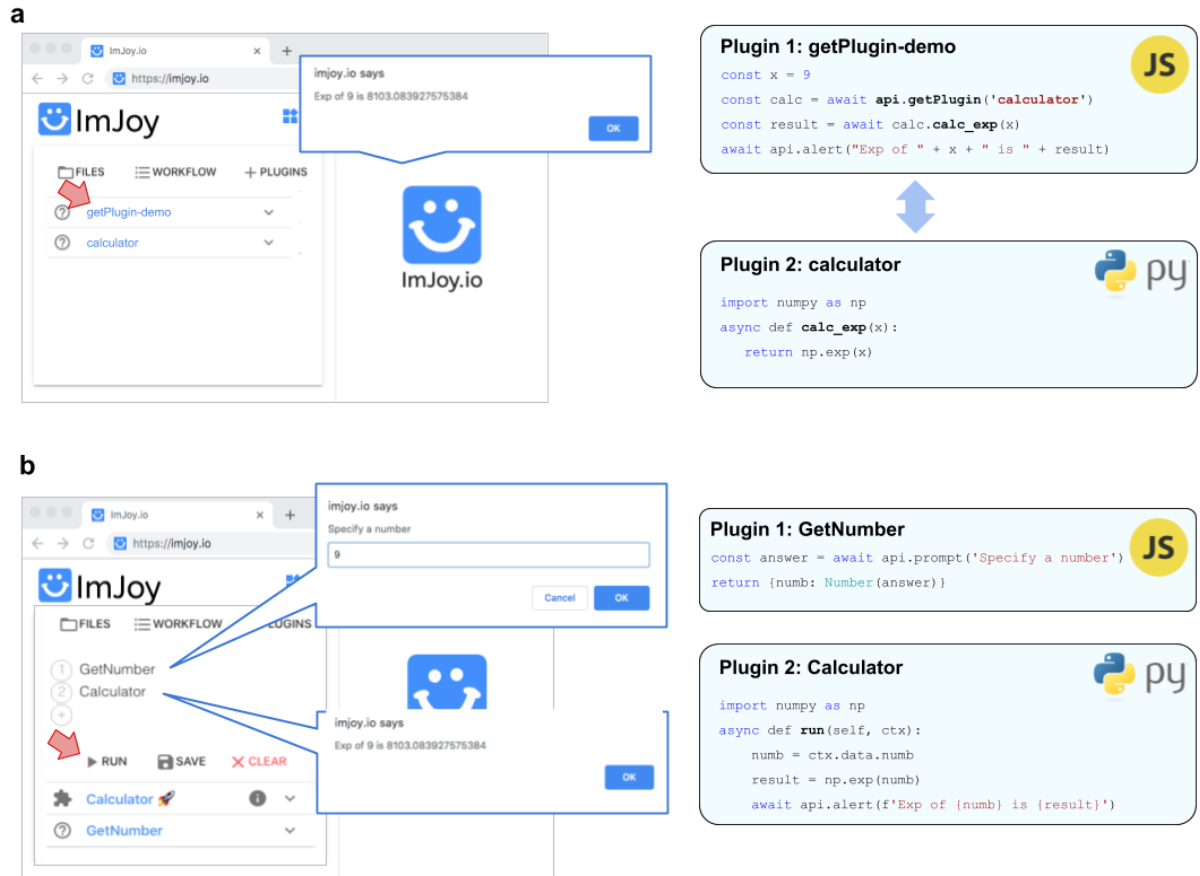

**Supplementary Figure 2: Plugin interactions in ImJoy**.

**a)** Workspaces. ImJoy allows to define different workspaces, each of which can contain different plugins. While keeping these plugins inside their own sandboxes or processes, the ImJoy API permits these plugins to call each other's API functions and exchange data within the same workspace. The same API can be used across programming languages or processes running on another computer. For example, a JavaScript plugin rendering a user interface can call functions of a Python plugin for data-processing and then display the results. In the example code snippets shown, a JavaScript plugin (`getPlugin-demo`) uses the ImJoy API function `api.getPlugin` to call the function `calc_exp` of a Python plugin (`calculator`) and calculate the exponential of the provided input. The result is then reported to the user. All API function calls are asynchronous to achieve concurrent execution of plugins. This is implemented by using `async/await` syntax with the function call as in the example plugin 2. **b)** Workflows. Plugins within a workspace can also be composed into a workflow, where multiple plugins are chained to perform more complex data analysis and visualization tasks. Note that these workflows can also be stored and shared with other users via a single URL. In the example shown, a JavaScript plugin asking the user to provide a number is executed first, then this number is used by a second plugin to perform calculations.

## Methods

Here, we provide a brief overview of the general ImJoy software design and its plugin system. For a more detailed and regularly updated description, we refer to the ImJoy documentation (https://imjoy.io/docs). We also list all example plugins with a brief description and installation links.

### ImJoy software design

ImJoy is developed as a minimal and flexible plugin-powered browser application built with modern web technologies (HTML5, JavaScript, WebAssembly). ImJoy provides a server-less solution with offline support, and runs on mobile devices. By design, ImJoy itself provides very minimal task-specific functionality. It offers a flexible plugin system with plugins allowing to provide user interfaces and perform computational tasks. ImJoy consists of three parts, and each can be extended with plugins (detailed below): (1) User Interface, where any existing web design library can be used to interact with the user; (2) Web Computational Backend to execute computational tasks directly inside web browsers; (3) Native Computational Backend (Plugin Engine) to perform more demanding computational tasks either locally or remotely.

We provide a detailed online documentation (https://imjoy.io/docs), which includes a user manual and description of all API functions with examples, dedicated demos to illustrate how to implement different plugin types, and a section with frequently asked questions (FAQ). Further, ImJoy is a community partner on image.sc, where questions for how to use ImJoy and process data can be asked in a dedicated forum (https://forum.image.sc/tags/imjoy).

ImJoy is open-source (MIT license) and entirely hosted on GitHub in several repositories for the browser app, the plugin engine, the desktop app, and different plugin repositories (more details on  https://imjoy.io).

**ImJoy web application**

The core of ImJoy is a set of static HTML/Javascript/CSS code which runs completely in the browser and can be served with any static web server. The official ImJoy website (https://imjoy.io) is hosted on GitHub and served through GitHub pages. When the user opens the ImJoy app with the browser, these static web pages will be loaded to the browser. No local data will be sent to the imJoy.io server. All data will remain local or exchanged only with the connected plugin engine. This so-called serverless design makes ImJoy suitable to work with sensitive data and offers maximum scalability. Lastly, this design guarantees that the user always obtains the latest ImJoy version upon relaunching the app.

The Desktop App (https://github.com/oeway/ImJoy-App) allows to use ImJoy locally without Internet connection, which is achieved by running a web server locally. The user can then access this locally served version of ImJoy (instead of https://imjoy.io ). While the ImJoy core can run offline entirely, some plugins may still require the Internet to fetch remote resources, e.g. javascript libraries or trained models for inference.

To keep the ImJoy application and plugins up to date, users are encouraged to use the online version. Note that this does not imply that the data will be sent to the server, since the ImJoy.io static web server does not accept any user data, and plugins process data either in the browser itself or through the local plugin engine.

**ImJoy plugin system**

Different plugin types are available to implement user interface, data display, or data processing in the browser or on the plugin engine. Developers can use a consistent API (application program interface) across different programming languages (JavaScript, Python), and `Asynchronous programming` to coordinate plugin processes that run in parallel. Plugins are individual html files, and can be directly developed in ImJoy with the built-in code editor. ImJoy provides isolated workspaces, where plugins for a given data processing task can be grouped.

Dependencies can be defined for each plugin, and will be automatically resolved by ImJoy. A plugin can run in an isolated virtual environment to avoid conflicts with other plugins. Further, plugin tags provide configurable modes for their dependency installation and execution, e.g. to specify operating system dependent versions of Tensorflow, or whether the plugin will use GPU or CPU for computation. This is especially important for modern deep learning libraries, which frequently require operating system and hardware specific dependencies.

Plugin files can be stored in any GitHub repository, which can then serve as a plugin repository. Plugin developers are thus clearly identified and can use ImJoy as a tool to facilitate usage of their code. Once deployed in the Github plugin repository, plugins can then be installed via a single web link. This will open ImJoy in the browser and allow the user to install the plugin. Several url parameters further allow to control how the plugin is installed and started. For example, a GitHub commit tag can be provided to control which version of the plugin is installed, thereby guaranteeing high reproducibility.

**Example plugins**

We designed these example plugins to demonstrate the versatility of ImJoy. For these demonstration plugins we deliberately chose to simplify the user experience, for example by hiding many hyperparameters for neural network training from the interface. However, these plugins can be easily adapted, extended or tailored for different use cases and for increased functionalities.

Below we list all example plugins mentioned in the paper. To install a plugin simply open the provided links starting with http://imjoy.io/#/app in a web browser. **Please note that ImJoy runs best on the latest version of Google Chrome (version 74.0+) or Firefox (version 66.0+).**

This link will open ImJoy and display the plugin installation dialog. After confirmation, the plugin will be installed in a dedicated workspace 'imjoy-examples'. The plugin will show up in the plugin list on the left side of the ImJoy interface. Installation progress is shown in the plugin-specific progress bar. More details can be found in the plugin log (accessible by clicking on the symbol left of the plugin name). Once a plugin is correctly installed, its name will turn blue and the plugin can be executed by pressing on it. Note that some plugins may install "helper" plugins which provide additional functionalities, e.g. a dashboard to monitor training progress. Such helper plugins will be shown in the plugin list with a grayed-out name.

Plugins running in the browser (indicated below with "Browser" in the title) can be used directly. The installation link of these plugins is configured such that they start automatically upon successful installation.

Plugins requiring the plugin engine are indicated below with "Engine" in the title. To run these plugins, the plugin engine has to be installed as described in the ImJoy user manual (https://imjoy.io/docs/#/user_manual?id=imjoy-app-and-plugin-engine). Once ImJoy is connected to this engine, the plugins can be assigned to it using the menu to the left of the plugin name. The specified workspace folder will then be created on the engine and the virtual environments and software dependencies of the plugin will be installed. More details are provided in the plugin log, which is accessible by pressing on the inverted exclamation mark next to the plugin name.

The provided deep learning plugins are built on open-source Python libraries with operating-system specific dependencies. For these plugins we therefore provide different "tags" for different operating system, that allow to control which libraries are installed. More details on the requirements to run the plugin and tested software environments are provided here: https://github.com/imjoy-team/example-plugins/blob/master/README.md

**Image Annotator plugin (Browser)**

Plugin installation link:

https://imjoy.io/#/app?w=imjoy-examples&plugin=oeway/ImJoy-Plugins:ImageAnnotator&start=ImageAnnotator&fullscreen=1

This browser plugin allows to manually outline and annotate different structures of interest in multi-channel images (TIFF or PNG). The plugin will open with a default image from the Human

Protein Atlas (HPA)[25], but other images can be loaded through the file menu if organized in the required folder structure. Different markers can be specified to annotate different structures, such as nuclei or microtubules. Multi-channel images can be shown as a composite image, or as individual channels. Annotations are stored in the geojson format and can be either exported as zip file or sent to a plugin engine for further processing, e.g. to create masks for training segmentation algorithms, as demonstrated in the DPNUnet segmentation plugin (see below).

Large data sets can be loaded into the plugin for annotation as folders. For more details on the required file organization we refer to the plugin documentation. Opening data with existing annotations, will result in an automatic display of these annotations.

**HPA-UMAP   (Browser)**

Plugin installation link:

https://imjoy.io/#/app?w=imjoy-examples&plugin=imjoy-team/example-plugins:HPA-UMAP&start=HPA-UMAP&fullscreen=1

This plugin allows to visualize results of a classification of protein localization patterns from microscope images of the entire HPA cell atlas[25] (v18, 67239 images).  Classification is performed with the deep-learning based winning method of the Human Protein Atlas Image Classification                                        Challenge                                         2018 (https://www.kaggle.com/c/human-protein-atlas-image-classification).  Plots shows a UMAP projection (Uniform Manifold Approximation and Projection) calculated from the last layer of the neural network (densenet121, dimensions: 1024). Each dot corresponds to one image, each color reflects a human annotated organelle localization pattern (e.g. red for nucleoplasmic localization, yellow for mitochondrial localization, and gray for mixed localization patterns). When hovering over UMAP plots, a thumbnail image for the currently selected data point is shown in the upper left corner. Clicking on a dot will show a larger view of this image together with information about the cell line, the gene corresponding to the protein, the annotated protein locations, and a link to the HPA website. The plugin also allows to search for a gene, cell line, or a localization pattern through a search box in the upper right corner, and display corresponding points in the UMAP.  The image slideshow allows to inspect the current image as an RGB composition, and also each of the four color channels (green, blue, yellow and red) individually. The green channel shows the protein of interest, while the other channels are reference markers targeting different cell organelles: blue - nucleus, yellow - endoplasmic reticulum, and red - microtubules.

**Skin-Lesion-Analyzer   (Browser)**

Plugin installation link:

[https://imjoy.io/#/app?w=imjoy-examples&plugin=imjoy-team/example-plugins:Skin-Lesion-Analyzer&start=Skin-Lesion-Analyzer](https://imjoy.io/#/app?w=imjoy-examples&plugin=imjoy-team/example-plugins:Skin-Lesion-Analyzer&start=Skin-Lesion-Analyzer)

**Disclaimer**: this plugin is intended for academic research only and not tested for medical diagnostics.

This plugin implements a deep learning methods that can classify an image of skin into 7 types of (potentially cancerous) skin lesions[3]. The model is trained with the deep learning library Keras (keras.io) and exported for use with Tensorflow.js (a Javascript implementation of the machine learning platform Tensorflow), and thus runs entirely in the browser. This allows the plugin to run on mobile devices such as smartphones. If the plugin is executed on a computer, the input image can be located on the hard drive; on a smartphone the image can be taken directly with the camera.

Code was ported from https://github.com/vbookshelf/Skin-Lesion-Analyzer.

**HPA-Classification   (Browser)**

Plugin installation link:

https://imjoy.io/#/app?w=imjoy-examples&plugin=imjoy-team/example-plugins:HPA-Classification&start=HPA-Classification&fullscreen=1

This plugin uses a deep model to predict the subcellular protein localization pattern from a provided image, based on annotated localization classes from the HPA cell atlas[25]. The model is a lightweight pre-trained neural network (ShuffleNet v2[29]) that enables fast prediction directly in the browser and won the special price of the Human Protein Atlas Image Classification Challenge 2018 (https://github.com/CellProfiling/HPA-Special-Prize). The plugin further displays a Class Activation Map[19], which indicates which parts of the image were used for predicting each class. The user can also select a new image from the HPA online database and predict the localization pattern.

**ANNA-PALM    (Engine)**

Plugin installation link:

https://imjoy.io/#/app?w=imjoy-examples&plugin=oeway/ImJoy-Plugins:ANNA-PALM

This plugin implements the ANNA-PALM method[6], which uses deep learning to reconstruct super-resolution views from sparse, rapidly acquired localization images and/or widefield images. The user can choose between training or prediction from a central launch pad. Training/prediction can be performed either on provided images of microtubules, or on user-supplied data in the csv format of ThunderSTORM. The launch pad also provides access to a detailed user manual, which is the recommended starting point for a new user.

More information on ANNA-PALM can be found at https://annapalm.pasteur.fr/#/. Code is ported from https://github.com/imodpasteur/ANNA-PALM

**CARE    (Engine)**

Plugin installation link:

https://imjoy.io/#/app?w=imjoy-examples&plugin=oeway/ImJoy-Plugins:CARE

This plugin implements an example of the deep learning-based CARE method (Content-aware image restoration) for denoising of 3D fluorescence microscopy images[5].

As for the ANNA-PALM plugin, the user can choose between training or prediction from a central launch pad and training can be performed either on provided example data (tribolium images) or new user-provided data. The launch pad also provides access to a detailed user manual, and is the recommended starting point for a new user.

Code is ported from  https://github.com/CSBDeep/CSBDeep/tree/master/examples/denoising3D

**DPNUnet    (Engine)**

Plugin installation link:

https://imjoy.io/#/app?w=imjoy-examples&plugin=https://raw.githubusercontent.com/oeway/DPNUnet-Segmentation/master/DPNUnet.imjoy.html

This plugin implements a deep learning model for segmenting nuclei in microscopy images and is based on the winning entry of the 2018 Kaggle Data Science Bowl on nuclear segmentation (https://www.kaggle.com/c/data-science-bowl-2018/discussion/5474). This method uses a Dual Path Network[26] (DPN) pretrained on ImageNet[27] with a U-net architecture[2]. Together with the Image Annotator plugin (above), the DPNUnet plugin offers a complete framework from image annotation to prediction. In practice, multiple plugins can be selected from a central launch pad: the annotation plugin ("annotate images") permits to browse and annotate training images by manually outlining nuclei; the mask generation plugin ("generate masks") can then be used to create masks based on these annotations; the training plugin ("train with data from the engine") then uses these masks as targets to re-train the DPNUnet model; finally, the trained model can be used for segmenting user-supplied images with the prediction plugin ("predict").

Code for segmentation is ported from https://github.com/selimsef/dsb2018_topcoders

Data for training the network is downloaded from

https://www.kaggle.com/c/data-science-bowl-2018

**DeepBindScan   (Engine)**

Plugin installation link:

https://imjoy.io/#/app?w=imjoy-examples&plugin=imjoy-team/example-plugins:DeepBindScan

This plugin uses a deep learning approach DeepBind to predict the binding affinities of transcription factors (TFs) and RNA binding proteins (RBPs) in different species to a user-provided DNA- or RNA- sequence[10]. It is implemented with Kipoi[28] (https://kipoi.org/), which provides ready-to-use trained models for genomics and a corresponding API. The user can specify a sequence, which will then be used as input to all available pre-trained DeepBind models (https://kipoi.org/groups/DeepBind/), each of which predicts the binding affinity of a single TF or RBP from a specific species. The models used require input sequences to be 101 base pairs long. Predicted binding affinities are shown in a vertical bar plot and updated dynamically. Final prediction results can also be exported as csv files.

Please note that for some organisms (e.g. homo sapiens), models for many proteins are provided, and the prediction will then take several minutes.